\documentclass[twoside]{IEEEtran}
\usepackage{amsmath}
\usepackage{amsfonts}
\usepackage{multirow}
\usepackage{lipsum}
\usepackage{here}
\usepackage{booktabs}
\usepackage{tabularray}

\ifCLASSINFOpdf
  \usepackage[pdftex]{graphicx}
  \DeclareGraphicsExtensions{.pdf,.jpeg,.png}
\else
\fi
\usepackage{scalerel}
\usepackage{tikz}
\usetikzlibrary{svg.path}

\definecolor{orcidlogocol}{HTML}{A6CE39}
\tikzset{
    orcidlogo/.pic={
        \fill[orcidlogocol] svg{M256,128c0,70.7-57.3,128-128,128C57.3,256,0,198.7,0,128C0,57.3,57.3,0,128,0C198.7,0,256,57.3,256,128z};
        \fill[white] svg{M86.3,186.2H70.9V79.1h15.4v48.4V186.2z}
        svg{M108.9,79.1h41.6c39.6,0,57,28.3,57,53.6c0,27.5-21.5,53.6-56.8,53.6h-41.8V79.1z M124.3,172.4h24.5c34.9,0,42.9-26.5,42.9-39.7c0-21.5-13.7-39.7-43.7-39.7h-23.7V172.4z}
        svg{M88.7,56.8c0,5.5-4.5,10.1-10.1,10.1c-5.6,0-10.1-4.6-10.1-10.1c0-5.6,4.5-10.1,10.1-10.1C84.2,46.7,88.7,51.3,88.7,56.8z};
    }
}

\newcommand\orcidicon[1]{\href{https://orcid.org/#1}{\mbox{\scalerel*{
                \begin{tikzpicture}[yscale=-1,transform shape]
                \pic{orcidlogo};
                \end{tikzpicture}
            }{|}}}}

\usepackage[bookmarks=false]{hyperref}

\begin{document}
%
\title{Mesh-Wise Prediction of Demographic Composition from Satellite Images Using Multi-Head Convolutional Neural Network}
%
%
%

\author{Yuta~Sato$^{\textsuperscript{\orcidicon{0009-0004-7052-7163}}}$
\thanks{}%
\thanks{Y. Sato is with the Department of Geography and Environment, London School of Economics and Political Science, Houghton Street, London WC2A 2AE, UK e-mail: (y.sato4@lse.ac.uk)}
}

\markboth{}{}
%



\maketitle

\begin{abstract}
Population aging is one of the most serious problems in certain countries. In order to implement its countermeasures, understanding its rapid progress is of urgency with a granular resolution. However, a detailed and rigorous survey with high frequency is not feasible due to the constraints of financial and human resources. Nowadays, Deep Learning is prevalent for pattern recognition with significant accuracy, with its application to remote sensing. This paper proposes a multi-head Convolutional Neural Network model with transfer learning from pre-trained ResNet50 for estimating mesh-wise demographics of Japan as one of the most aged countries in the world, with satellite images from Landsat-8/OLI and Suomi NPP/VIIRS-DNS as inputs and census demographics as labels. The trained model was performed on a testing dataset with a test score of at least 0.8914 in $\textbf{R}^2$ for all the demographic composition groups, and the estimated demographic composition was generated and visualised for 2022 as a non-census year.
\end{abstract}


%
\IEEEpeerreviewmaketitle

\section{Introduction}
%
%
%
%

\IEEEPARstart{W}{hile} the total population of human beings has increased radically, some countries and regions are being faced with unprecedented aging of society. As a forecast, 1 in 6 people in the world will be more than 60 years old, whose number will be doubled from 1 billion in 2020 to 2.1 billion in 2050 \cite{WHO_2022}. The radical change of demographic composition in countries and regions would affect the feasibility of social welfare and urban planning policies of governments such as pensions, medical care, transportation, and other physical and social infrastructure \cite{Bongaarts_2004, Alley_Liebig_Pynoos_Banerjee_Choi_2007}. In order to adjust to the transition of societies, obtaining up-to-date demographic composition is of urgent importance. As seen in the case of Italy \cite{Kertzer_White_Bernardi_Gabrielli_2009}, a low fertility rate stems from multiple factors including identity towards family, employability, and composition of family, but the importance of each factor highly varies depending on sub-regions. In terms of making data-driven decisions on the population aging with high geographic heterogeneity, it is not sufficient to only obtain the total population, but instead crucial to gain metrics of the demographic composition. One of the most accurate and detailed data sources for demographic composition is a census of each country. Although the granularity varies depending on countries and regions, a census provides precise attributes of residents in each tract or mesh at some timing of “census year”, such as population by age, number of households, and average income in the area. However, due to its thorough methodology and financial or administrative constraints, a census is usually conducted every several years, such as ten years in the U.S. and the U.K. \cite{Census_Gov_2022, Office_for_National_Statistics_2023}. Thus, it is not feasible to take place similar national surveys more frequently, such as on an annual basis, and approximation is required by estimation from other data sources.

Among others, Japan is now being suffered from the unprecedented population aging in the history of human beings. In estimation \cite{World_Bank_2023}, the ratio of people with ages over 64 years old occupied 30.0 percent of the whole population in Japan, which is the second highest among all the other countries next to Monaco. In parallel, the polarisation of the population is gaining momentum with Tokyo Metropolitan Area enjoying the population influx from the other regions since the 1970s \cite{MLITT_2012}. Under the circumstances, there are an increasing number of “marginal settlements”  (\emph{genkai shūraku}) in the rural areas, where lives became infeasible due to the lack of young labours and/or infrastructure and the residents would have to abandon their housings to migrate to the urban area \cite{Feldhoff_2012}. Considering the fast pace of population aging with an estimation of the population ratio over 64 years old as 38.8\% by 2050 \cite{Nakatani_2019}, it is crucial to understand the up-to-date demographic composition in a granular manner, from the perspective of urban and regional planning. Although the Statistics Bureau of Japan is providing the mesh-wise demographic composition as the result of the census, the interval of its provision is every five years \cite{Statistics_Bureau_of_Japan_2023b}. To fulfill the gap between the census years, precise estimation is of importance.

Nowadays, pattern recognition of non-linear relationships between different types of datasets is prevalent with potential application to problems in geography, due to the increase of data volume and new machine learning algorithms proposed \cite{Singleton_Arribas‐Bel_2019}. In particular, Deep Learning models as extensions of ANN have shown remarkable performances in the area of remote sensing, utilising satellite images (i.e. land-use classification \cite{Campos-Taberner_García-Haro_Martínez_Izquierdo-Verdiguier_Atzberger_Camps-Valls_Gilabert_2020}). Since some well-known satellites such as Landsat-8 \cite{USGS_2023a}, Sentinel-2 \cite{ESA_2023}, and Suomi NPP \cite{NOAA_2023} periodically observe the surface of the Earth globally, they constantly provide a rich amount of data source possibly enough to approximate the other socio-economic variables. When it comes to the estimation of population, previous studies have already tried to estimate the population density from satellite images using Convolutional Neural Network (CNN) such as in India \cite{Hu_Patel_Robert_Novosad_Asher_Tang_Burke_Lobell_Ermon_2019} and China \cite{Cheng_Wang_Feng_Yan_2021, Huang_Zhu_Zhang_Liu_Li_Zou_2021}. However, the estimation of demographic composition rather than the total population is understudied, especially in the context of population aging. This is mainly because the estimation of the population tended to focus on capturing the fast pace of urbanisation around cities in ‘developing’ countries, where rigid data source like census is radically not available. Moreover, some models from the previous studies relied on a complex mixture of multiple data sources, including POI data (i.e. \cite{Cheng_Wang_Feng_Yan_2021}). Although better prediction scores could be expected as the variety and amount of input data are increased, this approach implicitly lowers the generalisability of models because of data availability depending on the area of interest. Thus, it is important to develop prediction models by intentionally limiting the input data source only to those available commonly among countries and regions which are (and will be) suffering from population aging.

Being aware of the issues mentioned above, this paper proposes a CNN model of predicting the demographic composition of Japan on an annual basis, from the satellite images of Landsat-8/OLI and Suomi NPP/VIIRS-DNS, which are publicly available at any location on the Earth. Since the mesh-wise census demographic composition is available in Japan every five years, this paper employs demographic composition from the census conducted in 2015 and 2020 as labels for training, validation, and testing. At the end of the paper, a new dataset is synthesised from the satellite images in 2022, which is not a census year. Due to the public data availability in any country and region, this paper shall be significant in the sense that it has generalisability of granular estimation of demographic composition not only in Japan but also the other countries and regions.

\section{Related Works}

\subsection{Remote sensing for demographics estimation before Deep Learning}
In the field of remote sensing, satellite images have been utilised for a variety of purposes such as tracking desertification, aerosol pollution, and urbanisation \cite{Jenice_Raimond_2016}. Since its first launch in 1972, Landsat Missions operated by the U.S. Geological Survey (USGS) and National Aeronautics and Space Administration (NASA) have been employed as a major source of surface reflectance of the whole globe \cite{USGS_2023a}. Although the history is relatively shallower, Sentinel Mission from European Space Agency (ESA) followed Landsat as another source of satellite images since the first launch in 2014 with several applications including cropland monitoring, forest ecology, and urbanisation \cite{ESA_2023, Phiri_Simwanda_Salekin_Nyirenda_Murayama_Ranagalage_2020}. When it comes to demographics estimation, the urban population of Ohio in the U.S. was estimated from the impervious surface pixel value of Landsat-7 Enhanced Thematic Mapper (ETM+) images, by Ordinary Least Square (OLS) regression on a pixel basis and Spatial Autoregressive model (SAR) on a zonal basis \cite{Wu_Murray_2007}. However, as the authors stated, the accuracy of the zonal estimation was “unacceptable” due to the lack of capturing non-linearity in the pattern of pixels with a linear regression model.

While Landsat and Sentinel are observing the earth's surface reflectance in the daytime, other satellites focus on night-time light (NTL). With the global coverage, the Defence Meteorological Satellite Program’s Operational Linescan System (DMSP-OLS) by the U.S. Department of Defence and Suomi National Polar-Orbiting Partnership’s Visible Infrared Imaging Radiometer Suite Day-Night Band (Suomi-NPP/VIIRS-DNB) are ones of the major sources for NTL \cite{Barentine_Walczak_Gyuk_Tarr_Longcore_2021}. Previous studies pointed out that NTL could approximate socio-economic indicators such as land-use boundary \cite{Henderson_Yeh_Gong_Elvidge_Baugh_2003} and GDP \cite{Henderson_Storeygard_Weil_2012, Mellander_Lobo_Stolarick_Matheson_2015}. For the estimation of population, the correlations between NTL from DMSP-OLS and 1km mesh-wise population density were clarified in Hokkaido Prefecture, Japan by geographically weighted regression (GWR), with a prediction score of 0.8833 in $R^2$ \cite{Bagan_Yamagata_2015}. In another study \cite{Wang_Huang_Zhao_Hou_Zhang_Gu_2020}, 100m grid-wise population density was estimated in the extent of China using multiple sources including NTL from Suomi-NPP/VIIRS-DNB, Normalized Difference Vegetation Index (NDVI) from Moderate Resolution Imaging Spectroradiometer (MODIS), Digital Elevation Model (DEM), and the other POI data by Random Forest regression, whose accuracy score varied $R^2$ between 0.3 and 0.8 depending on provinces.

\subsection{Deep Learning and transfer learning with application to remote sensing}

The foundational concept of Artificial Neural Network (ANN) was initially proposed as a model of mimicking the human brain’s biological neurons as multi-layer perceptron (MLP) in the sense that input information propagates in a non-linear manner for pattern recognition of objects such as images, sounds, and sentences \cite{Rumelhart_Hinton_Williams_1986}. ANN introduced an activation function for each neuron, which enables the model to capture the non-linearity of input data sources. ANN is composed of multiple layers of neurons with weights, and each of the weights is trained through a chain of partial derivatives on a loss function called backpropagation. Although the concept was established earlier, it was computationally expensive at the moment when Graphics Processing Unit (GPU) was not adjusted for the computation of backpropagation \cite{Deng_2014}. With the advance of GPU, an ANN model called AlexNet was proposed for image recognition tasks with the introduction of a convolution layer, which captures patterns of neighbouring pixels efficiently by sharing parameters as filters \cite{Krizhevsky_Sutskever_Hinton_2017}. This 8-layer CNN model scored a remarkable top-5 classification error of 15.3\% on the ImageNet dataset consisting of 1.2 million 224 × 224 pixels with red-green-blue (RGB) colours with 1,000 image categories. AlexNet also introduced a new activation function called Rectified Linear Unit (ReLU) as $\sigma(x)=\max(0,x)$ for inputs $x$. Compared with previous activations functions such as sigmoid and tanh, ReLU performed better for capturing non-linearity in the CNN model. Since the success of AlexNet, a variety of CNN models have been proposed with increased accuracy and the number of convolution layers such as GoogLeNet with 22 layers \cite{Szegedy_Wei_Liu_Yangqing_Jia_Sermanet_Reed_Anguelov_Erhan_Vanhoucke_Rabinovich_2015} and VGGNet with up to 19 layers \cite{Simonyan_Zisserman_2015}. However, the increase of layer numbers did not simply reflect on better prediction, as the gradient in backpropagation would be unstable to be explosive or vanished. To solve this problem and make the layers “deeper”, ResNet was proposed by adopting residual layers which introduced skip connections from previous convolution layers \cite{He_Zhang_Ren_Sun_2016}. With this improvement, ResNet could augment layers up to 152 layers, with 3.57\% top-5 classification error on the ImageNet dataset, which outperformed the error of human beings with 5.1\% \cite{Russakovsky_Deng_Su_Krause_Satheesh_Ma_Huang_Karpathy_Khosla_Bernstein_et_al_2015}.

As the accuracy of CNN models was improved, well-known pre-trained CNN models were fine-tuned and re-utilised for multiple domains other than the image classification of the ImageNet dataset, namely transfer learning \cite{Zhuang_Qi_Duan_Xi_Zhu_Zhu_Xiong_He_2021}. While default CNN models require a huge amount of datasets (i.e. 1,000 categories classification of 1.2 million images from ImageNet), transferred ones can achieve a considerable level of prediction with fewer samples if the task domain is similar. One of the simplest ways of transfer learning is to append or modify the last layer of models with holding the others the same at the beginning of training. For instance, ImageNet-based CNN models including AlexNet, GoogLeNet, and VGGNet could be transferred via fine-tuning into two medical image detection tasks from CT-scanned slices \cite{Shin_Roth_Gao_Lu_Xu_Nogues_Yao_Mollura_Summers_2016}.

In the field of remote sensing, CNN models have been mainly used for image segmentation tasks, where the pixel-wise classification output size is the same as the one from input images, such as cloud detection \cite{Yang_Guo_Yue_Liu_Hu_Li_2019}, deforestation monitoring \cite{de_Bem_de_Carvalho_Junior_Fontes_Guimarães_Trancoso_Gomes_2020}, and land-use classification \cite{Ulmas_Liiv_2020, Solórzano_Mas_Gao_Gallardo-Cruz_2021}. In the case of demographics estimation, previous studies proposed CNN models for regression tasks of population density. As an example, ImageNet-pre-trained CNN models including VGGNet and ResNet were employed for estimating the total population in 30 arc-second grids (approximately 1 km on the equator) in the U.S., using Sentinel-2 images in RGB and near-infrared (NIR) \cite{Huang_Zhu_Zhang_Liu_Li_Zou_2021}. The test error of logged population in the Metro Dallas area scored $R^2$ of 0.875 and 0.906 for VGGNet and ResNet, respectively. Others developed an original 6-layer CNN model with feature extraction of MLP estimated the 100m mesh-wise population density in Shenzhen, China, using multiple data sources including NTL from VIIRS-DNS and NDVI from the Chinese Academy of Sciences, and social sensing POI data \cite{Cheng_Wang_Feng_Yan_2021}. Although daytime multispectral images like Landsat and Sentinel were not used, the prediction on the testing dataset counted 0.77 in $R^2$. While these previous studies showed the success of CNN models in the task of population estimation, its more detailed breakdown into the demographic composition is understudied. Also, some of the data sources employed included local dataset which was only available in specific countries or regions, causing the limitation in generalisability of trained models into other geographic locations.

\section{Datasets and study areas}

To proceed with the development of a CNN model for estimating demographic composition in Japan, this paper employed Landsat-8/OLI Collection 2 Level-2 \cite{USGS_2023b} and Suomi NPP/VIIRS-DNS \cite{NOAA_2023} as inputs, and mesh-wise population census data provided by Statistics Bureau of Japan via e-Stat database \cite{eStat_2023} as ground truth labels. Samples were obtained from 2015 and 2020 with labels of demographic composition and 2022 without labels as a non-census year for generating estimated labels. The geographic extent of the samples was defined as the whole meshes available in Japan either from the census of 2015 or 2020. Each of the mesh cells was geographically defined as “Basic Grid Square” with a geodesic system of WGS 84, which spans 30 arc-seconds and 45 arc-seconds (approximately 1km square in the territory of Japan) for latitude and longitude, respectively \cite{Statistics_Bureau_of_Japan_2023a}. While the population census provides the number of residents in the mesh cell by age as of October 1st in the census year, aggregated number of three demographic composition groups of residents below 14 years old, between 15 and 64 years old, and over 64 years old were used as labels for a CNN model. This paper utilised satellite images both from daytime (Landsat-8/OLI) and night-time (Suomi NPP/VIIRS-DNS) so that the input could include more detailed information about land-use and human activities. Assumedly, the existence of only either building structures or NTL may not necessarily indicate the settlement of people (i.e. abandoned housings in rural areas; 24-hour factories in industrial areas). Thus, stacking multiple images as a single tensor could support a CNN model to recognise the demographic composition in the area from multiple aspects in the inputs. Landsat-8/OLI consists of nine multispectral images depending on the wavelength of light. Out of the nine bands, band 4, 3, and 2 were used as RGB images of natural colour. In addition to it, band 7, 6, and 5 were also used for another RGB image as false colour, so that wavelengths out of human eyes could also be recognised by a CNN model. Moreover, three synthesised indices were appended as the third RGB image from the original Landsat-8/OLI multispectral images as follows; NDVI, Normalized Difference Built-up Index (NDBI), and Normalized Difference Water Index (NDWI). While NDVI as an index for vegetation was already used for population estimation (i.e. \cite{Wang_Huang_Zhao_Hou_Zhang_Gu_2020}), this paper additionally employed the indices for buildings and water to explicitly delineate the different characteristics of lands. By referring to \cite{Kaplan_Avdan_Avdan_2018, Özelkan_2020}, these three indices were calculated as follows:

\begin{equation}
\text{NDBI} = \frac{\text{MIR} - \text{NIR}}{\text{MIR} + \text{NIR}} = \frac{\text{Band 6} - \text{Band 5}}{\text{Band 6} + \text{Band 5}}
\end{equation}

\begin{equation}
\text{NDVI} = \frac{\text{NIR} - \text{Red}}{\text{NIR} + \text{Red}} = \frac{\text{Band 5} - \text{Band 4}}{\text{Band 5} + \text{Band 4}}
\end{equation}

\begin{equation}
\text{NDWI} = \frac{\text{Green} - \text{NIR}}{\text{Green} + \text{NIR}} = \frac{\text{Band 3} - \text{Band 5}}{\text{Band 3} + \text{Band 5}}
\end{equation}

where NIR and MIR stand for near-infra-red and middle-infra-red, respectively.

\begin{figure*}[h]
\centering
\includegraphics[width=0.8\textwidth]{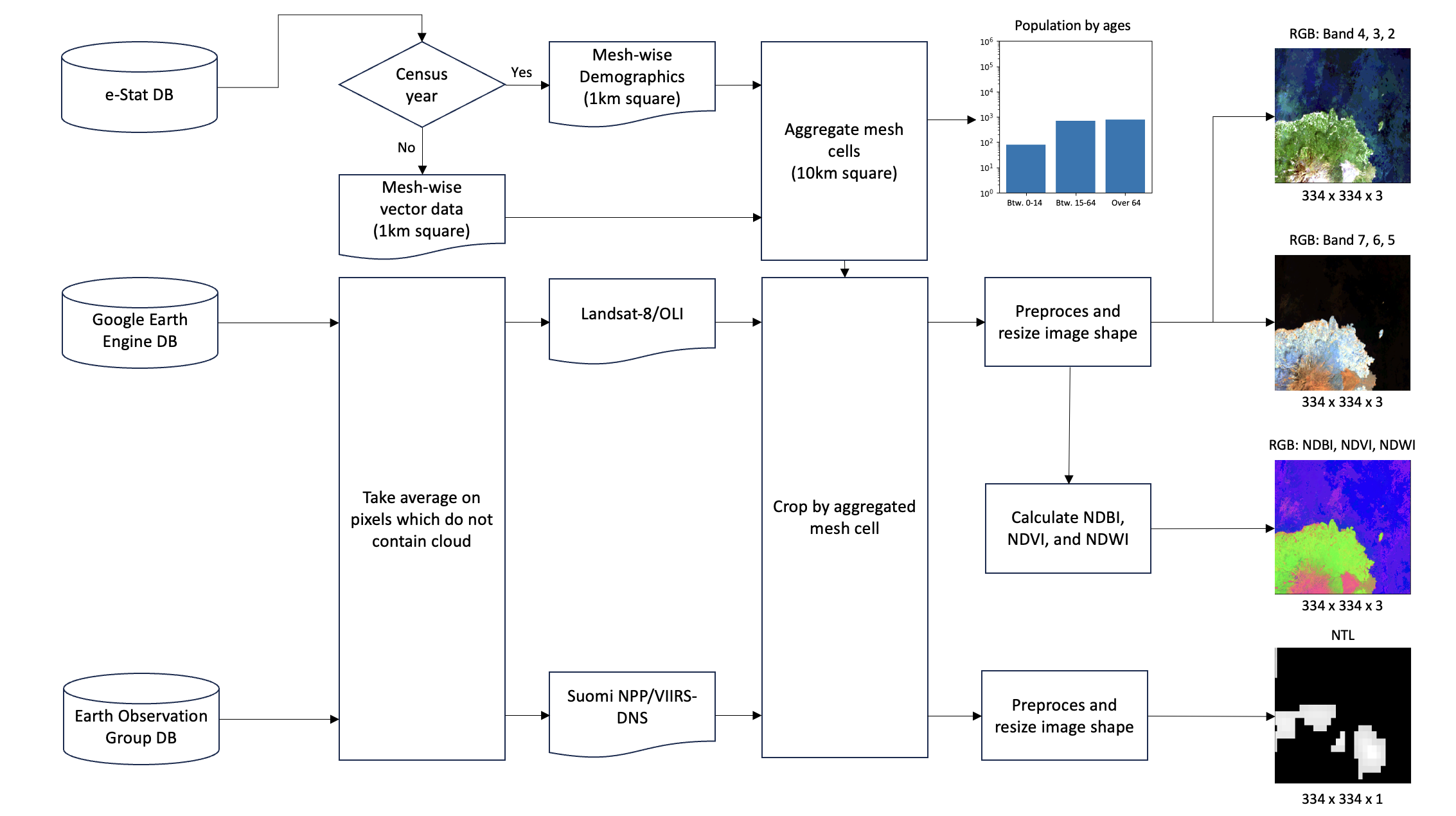}
\caption{Flow chart of data processing per one single sample.}
\label{figure_process}
\end{figure*}

\begin{figure*}[h]
\centering
\includegraphics[width=0.8\textwidth]{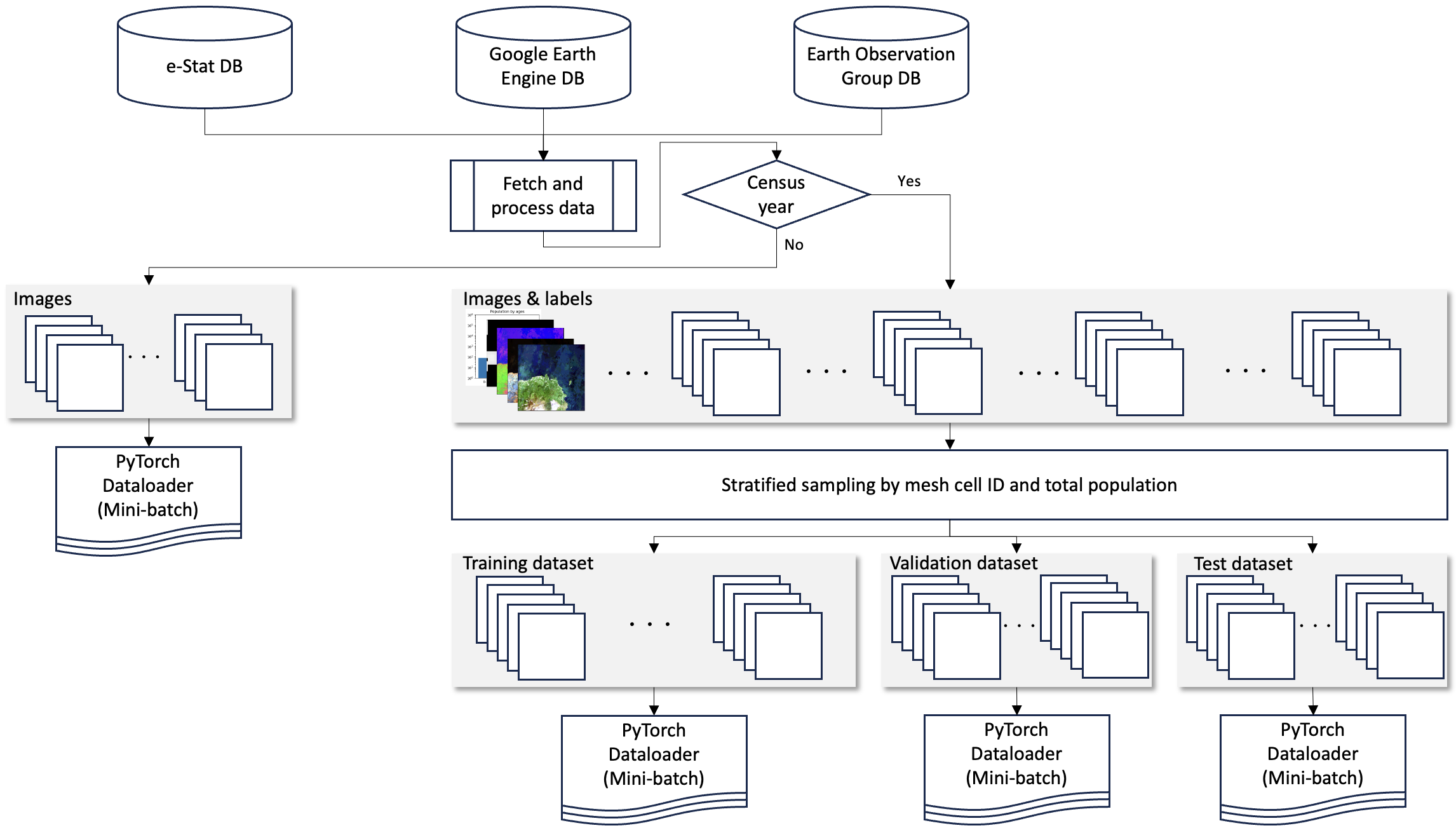}
\caption{Flow chart of data splitting for the supervised learning framework.}
\label{figure_split}
\end{figure*}

\begin{figure*}[h]
\centering
\includegraphics[width=\textwidth]{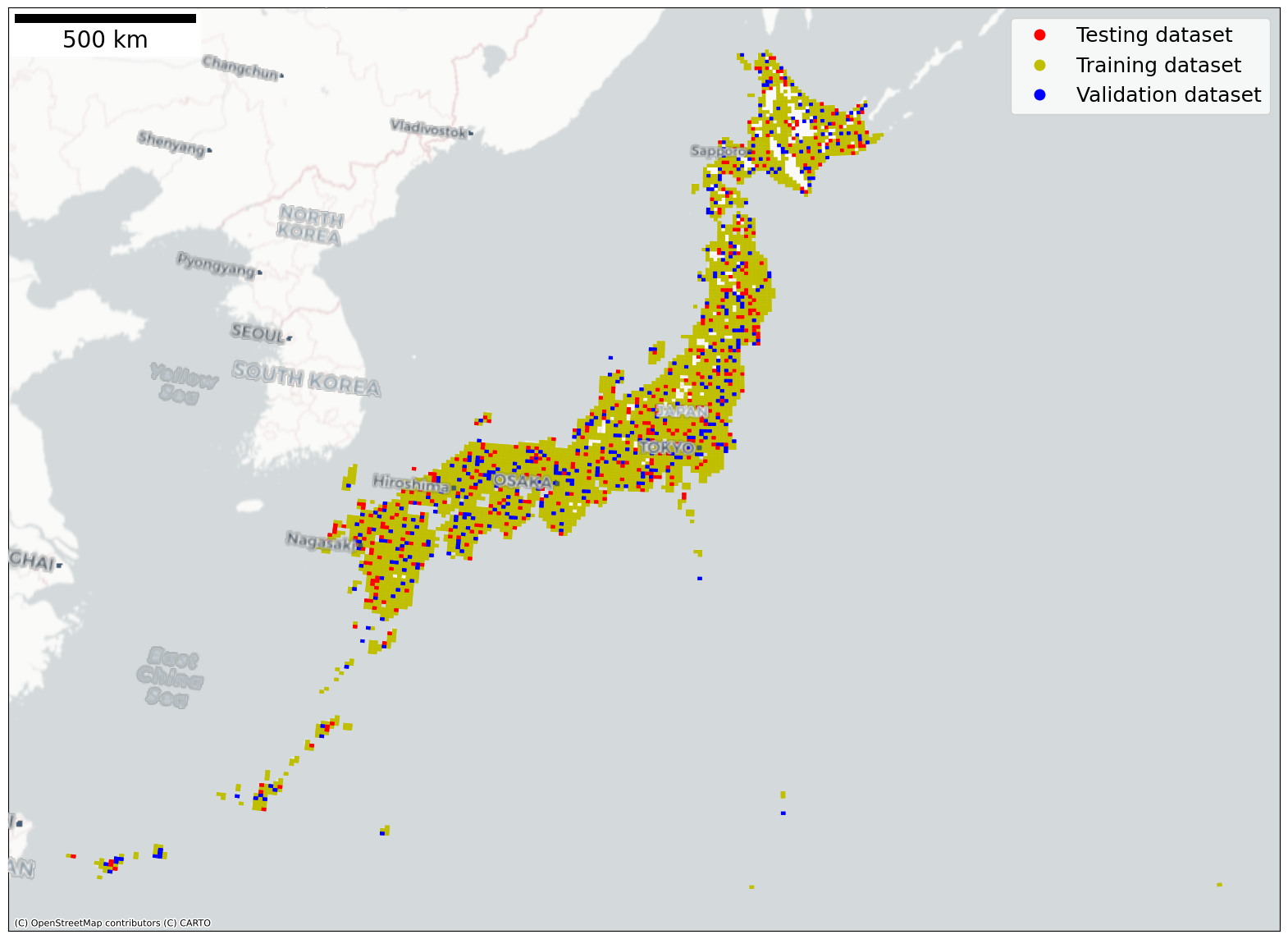}
\caption{Spatial distribution of training, validation, and test dataset by two census years. JGD2011 was applied for the coordinate reference system. The map is based on OpenStreetMap.}
\label{figure_spatial_distribution}
\end{figure*}

\begin{table*}[h]
\centering
\renewcommand{\arraystretch}{1.3}
\caption{A summary of descriptive statistics of samples by dataset and sample year.}
\label{table_summary}
\begin{tabular}{ccc|ccccccccc}
\hline
                                    &                          &       & \multicolumn{3}{c}{\begin{tabular}[c]{@{}c@{}}Population of ages between\\ 0 and 14 years old\end{tabular}} & \multicolumn{3}{c}{\begin{tabular}[c]{@{}c@{}}Population of ages between\\ 15 and 64 years old\end{tabular}} & \multicolumn{3}{c}{\begin{tabular}[c]{@{}c@{}}Population of ages\\ over 64 years old\end{tabular}} \\ \hline
Dataset                             & Year                     & Count & Mean                               & Median                            & Std.                               & Mean                                & Median                            & Std.                               & Mean                              & Median                            & Std.                               \\ \hline
\multirow{2}{*}{Training}           & 2015                     & 3367  & 3778.3                             & 311.0                             & 13212.7                            & 18118.2                             & 1665.0                            & 69416.7                            & 7957.6                            & 1315.0                            & 25409.9                            \\
                                    & 2020                     & 3357  & 3571.8                             & 255.0                             & 13176.6                            & 17382.7                             & 1391.0                            & 69913.6                            & 8418.0                            & 1304.0                            & 26558.3                            \\
\multirow{2}{*}{Validation}         & 2015                     & 419   & 4038.5                             & 304.0                             & 13989.0                            & 19945.3                             & 1620.0                            & 75072.0                            & 8317.7                            & 1391.0                            & 26142.8                            \\
                                    & 2020                     & 421   & 3815.5                             & 246.0                             & 13942.7                            & 19105.3                             & 1332.0                            & 75216.7                            & 8783.5                            & 1360.0                            & 27536.5                            \\
\multirow{2}{*}{Testing}            & 2015                     & 421   & 3499.5                             & 324.0                             & 10005.5                            & 16454.9                             & 1734.0                            & 48012.7                            & 7570.3                            & 1359.0                            & 19495.1                            \\
                                    & 2020                     & 418   & 3250.5                             & 266.0                             & 9453.2                             & 15611.4                             & 1425.5                            & 46691.9                            & 8083.2                            & 1334.5                            & 20995.1                            \\
\multicolumn{1}{l}{Non-census year} & \multicolumn{1}{l}{2022} & 4225  & -                                  & -                                 & -                                  & -                                   & -                                 & -                                  & -                                 & -                                 & -                                  \\ \hline
\end{tabular}
\end{table*}

\begin{figure*}[h]
\centering
\includegraphics[width=4in]{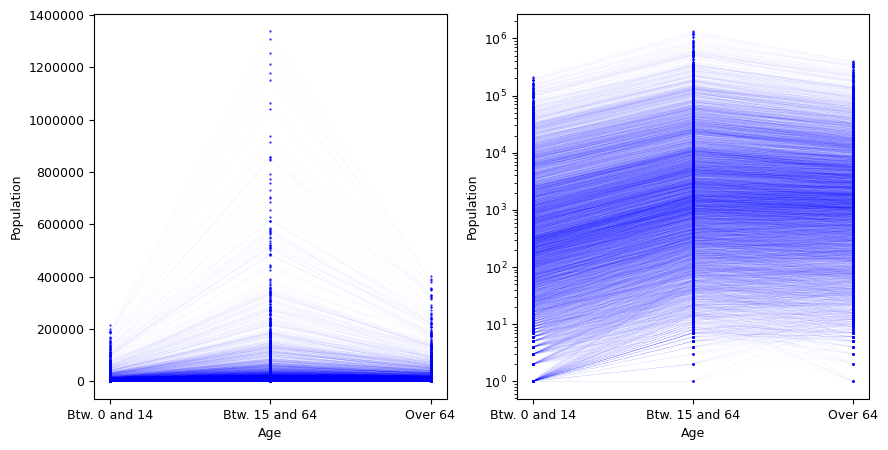}
\caption{Distribution of population in all the samples from the training, validation, and testing dataset by demographic composition groups. Scales on the y-axis are original and logarithm with base 10 for the left and right figure, respectively. The blue lines across ages indicate that the values were from the same sample in a mesh cell either in 2015 or 2020. }
\label{figure_pop}
\end{figure*}

\begin{figure*}[h]
\centering
\includegraphics[width=0.8\textwidth]{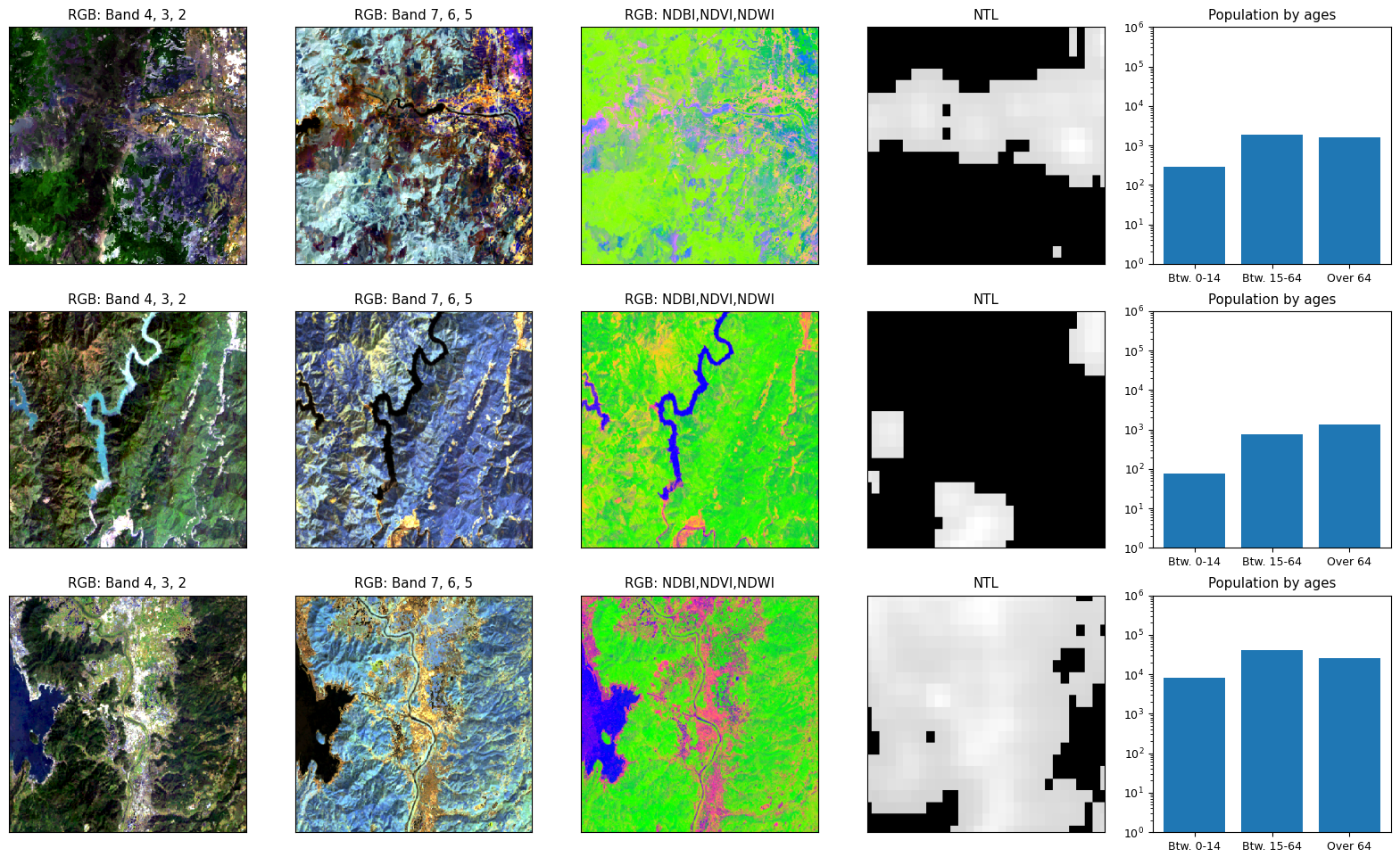}
\caption{Images and labels from three random samples from the training dataset.}
\label{figure_samples}
\end{figure*}

Figure~\ref{figure_process} illustrates the overview of processing images from the three data sources for a CNN model. Since satellite images are periodically collected from the space as snapshots of the Earth, it is inevitable to encounter cloud coverage in the images. To avoid biases from cloud coverage, this paper employed adjusted images where each pixel is an annual average of those without clouds. For Landsat-8/OLI, Google Earth Engine API was used for obtaining the adjusted images of “USGS Landsat 8 Level 2, Collection 2, Tier 1” \cite{Google_2023}. In the case of Suomi NPP/VIIRS-DNS, adjusted raster files were acquired from Annual VNL V2.1 maintained by the Earth Observation Group \cite{Elvidge_Zhizhin_Hsu_Baugh_2013, Elvidge_Zhizhin_Ghosh_Hsu_Taneja_2021, Earth_Observation_Group_2023}. The spatial resolutions of Landsat-8/OLI and Suomi NPP/VIIRS-DNS are approximately 30m and 750m, respectively. To fulfill the gap, the pixels of Suomi NPP/VIIRS-DNS were adjusted to those from Landsat-8/OLI. Due to the coarse spatial resolution of satellite images for the size of original meshes from the census, mesh cells were firstly aggregated from 30 arc-seconds $\times$ 45 arc-seconds (1km square) to 5 arc-minutes $\times$ 7 arc-minutes and 30 arc-seconds (10km square). After the cloud coverage removal by annual averaging, two types of satellite images were cropped by the aggregated mesh cells. The cropped images were pre-processed for resizing the shape into 334 $\times$ 334 pixels square so that the following CNN model can deal with the images as inputs with 30m spatial resolution for the 10km square extent. In the pre-processing, the values of Suomi NPP/VIIRS-DNS were re-scaled with a natural logarithm. The output of this flow was the stacked images of Band 2 to 7 from Landsat-8/OLI, their synthesised three indices (NDBI, NDVI, NDWI), and the NTL from Suomi NPP/VIIRS-DNS.

Based on the processing per sample explained in Figure~\ref{figure_process}, Figure~\ref{figure_split} illustrates the flow of sample management as a framework of a supervised learning algorithm. In order to train a CNN model for the task of demographic composition estimation, all the processed images and labels from the census year of 2015 and 2020 were split into those for training, validation, and testing. In order to ensure even distribution among the dataset groups, stratified sampling was employed by mesh cell ID and total population. In other words, unique values of strings for mesh cell IDs were extracted from all the samples, and averaged values of the total population between 2015 and 2020 were obtained. It is noteworthy to mention that there was no geographical duplication between the training, validation, and testing dataset since it is already rare to observe a radical urbanisation process in Japan and geographical duplication might cause data leakage problems \cite{Kaufman_Rosset_Perlich_2011} in the supervised learning framework. The mesh cell IDs were assigned to 10 partitions by ten deciles of the averaged total population in the histogram. Out of the 10 partitions, samples were evenly split into training, validation, and testing with a ratio of 8:1:1. The split samples were imported to PyTorch Dataloader objects \cite{Paszke_Gross_Massa_Lerer_Bradbury_Chanan_Killeen_Lin_Gimelshein_Antiga_et_al_2019} for mini-batch learning of a CNN model. Images from 2022 were directly imported to PyTorch Dataloader objects without labels. Figure~\ref{figure_spatial_distribution} shows the spatial distribution of samples from the training, validation, and testing dataset from the years 2015 and 2020. For the mesh cells where there was no data was omitted from the output since it does not necessarily mean the non-existence of residents, but also implies the lack of conducting a population census (i.e. Japanese government is claiming the sovereignty of the Kuril Islands without conducting a population census since they are under the effective control of Russian government). Table~\ref{table_summary} describes the summary statistics of samples by the type of dataset and sample year. As seen on the left in Figure~\ref{figure_pop}, the distribution of the population is heavily skewed. On the other hand, the logarithm with base 10 was applied for the original number of the population, which is shown on the right in Figure~\ref{figure_pop}. Thus, for the labels of the CNN model, this paper employs the logarithm of the original population with base 10, However, some samples contain 0 for certain demographic composition groups, where the logarithm is not defined. To avoid errors in the following computation, 0 was replaced with 1 ($10^0$) in advance. Figure~\ref{figure_samples} shows three randomly sampled images and labels. As seen in the figure, the images with RGB of Band 4, 3, and 2 were close to natural colour, while those with Band 7, 6, 5 and NDBI, NDVI, and NDWI were expressed in false colour. The combination of NDBI, NDVI, and NDWI could explicitly delineate water, forest, and others. NTL from Suomi NPP/VIIRS-DNS highlighted the location of human activities at night.

%

\section{Methodology}

Let the CNN model $f:= \mathcal{X} \to \mathcal{Y}$, where $\mathcal{X}:= \{\mathbf{X}_i | \mathbf{X}_i \in \mathbb{R}^{334 \times 334 \times 12} \}$ a feature tensor space consisting of four types of RGB-based images and $\mathcal{Y}:= \{\mathbf{y}_i | \mathbf{y}_i \in \mathbb{R}^{3} \}$ a label vector space consisting of three number of demographic composition groups. Then, the predicted demographic composition vector $\hat{\mathbf{y}}_{it} \in \mathcal{Y}$ is expressed as following:

\begin{equation}
\hat{\mathbf{y}}_{it} = f(\mathbf{X}_i; \boldsymbol \Theta_t)
\end{equation}

where $\boldsymbol \Theta_t$ is an element of $\boldsymbol \Theta$ as a set of layer weights in $f$ such that $\boldsymbol \Theta = \{\boldsymbol \Theta_t \}$ and $t \in \mathbb{N}$ denotes the order of training. The CNN model $f$ was trained through the gradient descent method as backpropagation such that

\begin{equation}
\boldsymbol \theta^{l}_{t} = \boldsymbol \theta^{l}_{t-1} - \text{Optim}(\alpha, \nabla \boldsymbol \theta^{l}_{t-1})
\end{equation}

\begin{equation}
\nabla \boldsymbol \theta^{l}_{t-1} = \frac{\partial \mathcal{L}(\hat{\mathbf{y}}_{it-1}, \mathbf{y}_i)}{\partial \boldsymbol \theta^{l}_{t-1}}
\end{equation}

where $\boldsymbol \theta^{l}_{t} \in  \boldsymbol \Theta_{t}$ and $\text{Optim}(\alpha, \nabla \theta^{l}_{t})$ denote the $l$-th layer of $\boldsymbol \Theta_{t}$ and the optimiser with the learning rate $\alpha \in \mathbb{R}^{+}$ and gradient $\nabla \theta^{l}_{t}$, respectively. For the loss function $\mathcal{L}(\hat{\mathbf{y}}_{it}, \mathbf{y}_i)$, this paper adopts the formula of following:

\begin{equation}
\mathcal{L}(\hat{\mathbf{y}}_{it}, \mathbf{y}_i) = \frac{1}{N} \sum_{i}^{N} \lvert \lvert \hat{\mathbf{y}}_{it} - \mathbf{y}_i \rvert \rvert 
\end{equation}

where $N \in \mathbb{N}$ denotes the number of samples in a dataset. For the optimiser, Adam \cite{Kingma_Ba_2017} was applied with following:

\begin{equation}
\text{Optim}(\alpha, \nabla \boldsymbol \theta^{l}_{t}; \beta_1, \beta_2) = \alpha \frac{\hat{m}_t}{\sqrt{\hat{v}_t + \epsilon}} \nabla \boldsymbol \theta^{l}_{t-1}
\end{equation}

\begin{equation}
m_{t} = \beta_1 m_{t-1} - (1 - \beta_1) \nabla \boldsymbol \theta^{l}_{t-1}
\end{equation}

\begin{equation}
v_{t} = \beta_2 v_{t-1} - (1 - \beta_2) (\nabla \boldsymbol \theta^{l}_{t-1})^2
\end{equation}

\begin{equation}
\hat{m}_{t} = \frac{m_{t}}{1 - \beta_1^t}
\end{equation}

\begin{equation}
\hat{v}_{t} = \frac{v_{t}}{1 - \beta_2^t}
\end{equation}

where $\beta_1$ and $\beta_2$ denote the hyperparameters of the first and second moment vectors. After every training of all samples in the training dataset as one epoch, the trained CNN model $f$ was validated with samples in the validation dataset by calculating $\mathcal{L}$. This one cycle of training and validation was iterated over 250 epochs, and a model with the least validation error was selected as the best model in terms of generalisability. The chosen best model was tested by samples in the testing dataset and $R^2_{(g)}$ was calculated for all the demographic composition groups $g$ as follows:

\begin{equation}
R^{2}_{(g)} = \frac{\sum (\hat{y}_i^{(g)} - \bar{y}_i^{(g)})^2}{\sum (y_i^{(g)} - \bar{y}_i^{(g)})^2}    
\end{equation}

\begin{figure*}[h]
\centering
\includegraphics[width=0.8\textwidth]{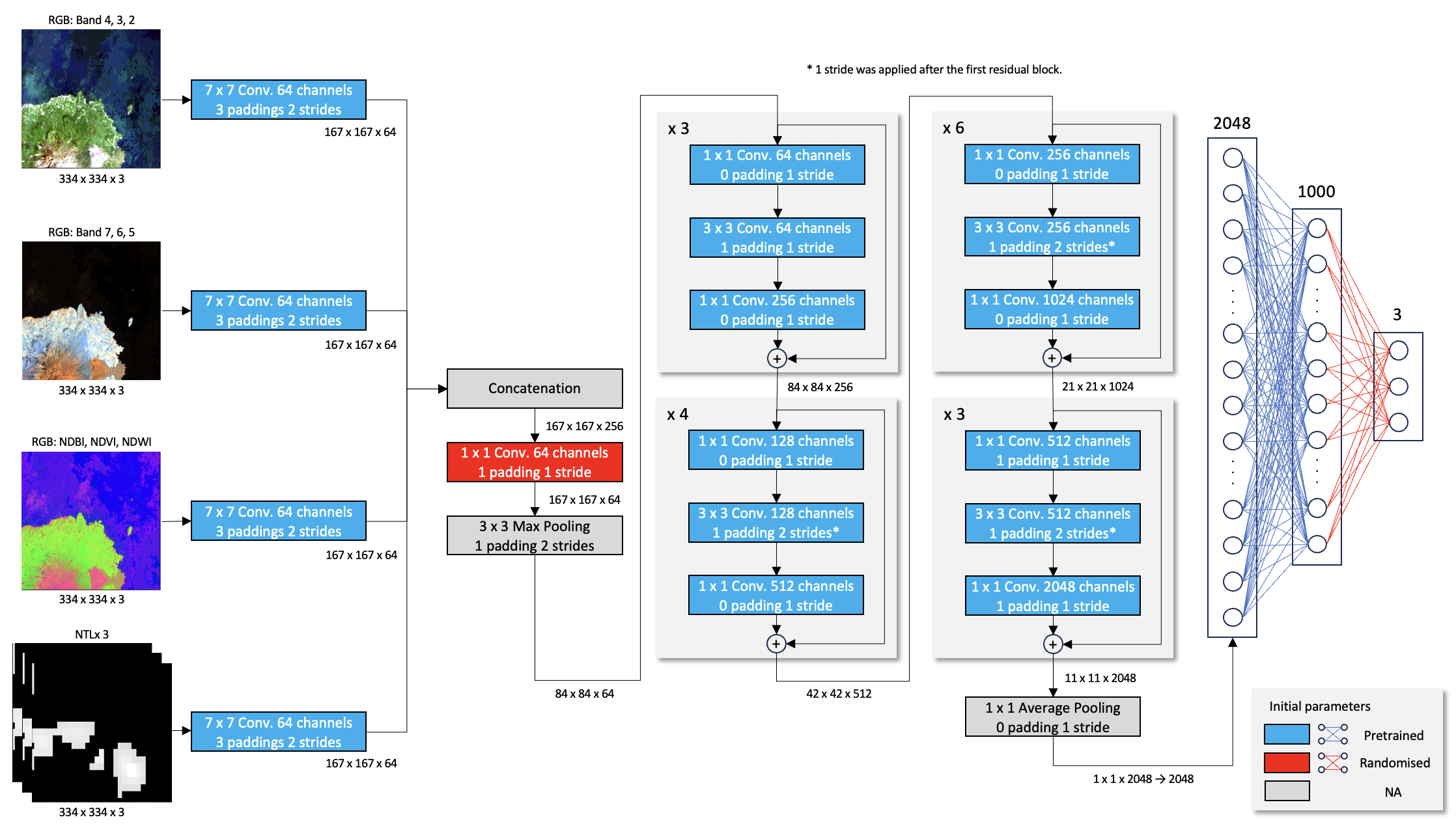}
\caption{Architecture of ResNet50-based multi-head CNN model.}
\label{figure_architecture}
\end{figure*}

Figure \ref{figure_architecture} displays the architecture of the CNN model $f$. The model employed multi-head CNN based on ResNet50 \cite{He_Zhang_Ren_Sun_2016}, so that multi-layered images could be processed as multiple RGB-coloured tensors. Four images were set as inputs of RGB-coloured tensors, where NTL was duplicately stacked for each colour. These inputs were passed through the first convolution layer and concatenated along the dimension of channels. The concatenated tensor was convoluted with 1 $\times$ 1 filter and aggregated by max pooling layer for down-sizing, which only keeps a maximum value in the window. The pooled tensor was further convoluted 48 times with skip connections. The output from all the convolutions was pooled by 1 $\times$ 1 average pooling layer as an operation of taking an average in the window, leading to 1 $\times$ 1 $\times$ 2048 tensor. The tensor was flattened to a 2048-dimension vector and passed through two fully-connected layers with the final output of a 3-dimension vector $\hat{\mathbf{y}}_{it}$, estimating the demographic composition in the sample. The output size of each convolution and pooling layer was described in Figure \ref{figure_architecture} by the formula of following:

\begin{equation}
n_{\textrm{out}} = \lfloor \frac{n_{\textrm{in}} + 2p - k}{s} \rfloor + 1
\end{equation}

where $n_{\textrm{in}}$ and $ n_{\textrm{out}}$ denote the size of input and output, respectively. $p$, $k$, and $s$ stand for the size of zero-padding, filter, and strides, respectively.  $\lfloor . \rfloor$ operation indicates obtaining an integer number from the float number. For instance, the output size of the first convolution for Landsat-8/OLI Band 4,3, and 2 was $\lfloor \frac{334 + 2*3 - 7}{2} \rfloor + 1 = 167$. As shown in Figure~\ref{figure_architecture}, all the parameters of this CNN model were transferred from ResNet50 being pre-trained by the ImageNet dataset, except the second convolution layer and the last fully-connected layer. Moreover, batch normalisation was implemented after each of the convolution layers, where each batch in the forward propagation was normalised with the average and variance, so that the training process could be stabilised \cite{Ioffe_Szegedy_2015}. $\text{ReLU}$ was used as an activation function for all the convolution layers after batch normalisation. In addition to batch normalisation, dropout \cite{Krizhevsky_Sutskever_Hinton_2017} was implemented for avoiding overfitting. Dropout is an operation for removing a certain amount of neurons within a layer during training so that the CNN model would be robust to the heterogeneity in the dataset as if it behaves as ensemble learning. However, as a previous study reported \cite{Li_Chen_Hu_Yang_2019}, the co-existence of dropout and batch normalisation in the convolution layer would worsen the performance of CNN models due to variance shifts. Thus, dropout was applied only to the last two fully-connected layers, where batch normalisation was not implemented in the original ResNet50. Since all the labels were biased to $10^0$ and their values were always non-negative, the last layer was also passed through $\text{ReLU}$. For the hyperparameters of this machine learning framework, the learning rate $\alpha$ and coefficients of the first and second moment $\beta_1$ and $\beta_2$ for Adam were set as 0.001, 0.9, and 0.999, respectively. The ratio of dropout was set as 0.25 for both of the two fully-connected layers. As an environment of model training, AWS Sagemaker Studio was used with an instance type of g4dn.xlarge consisting of 4 vCPUs, 16 GB memory, and 1 GPU of NVIDIA Tesla T4 \cite{Amazon_2023}.

\section{Result}

\begin{figure*}[!t]
\centering
\includegraphics[width=0.8\textwidth]{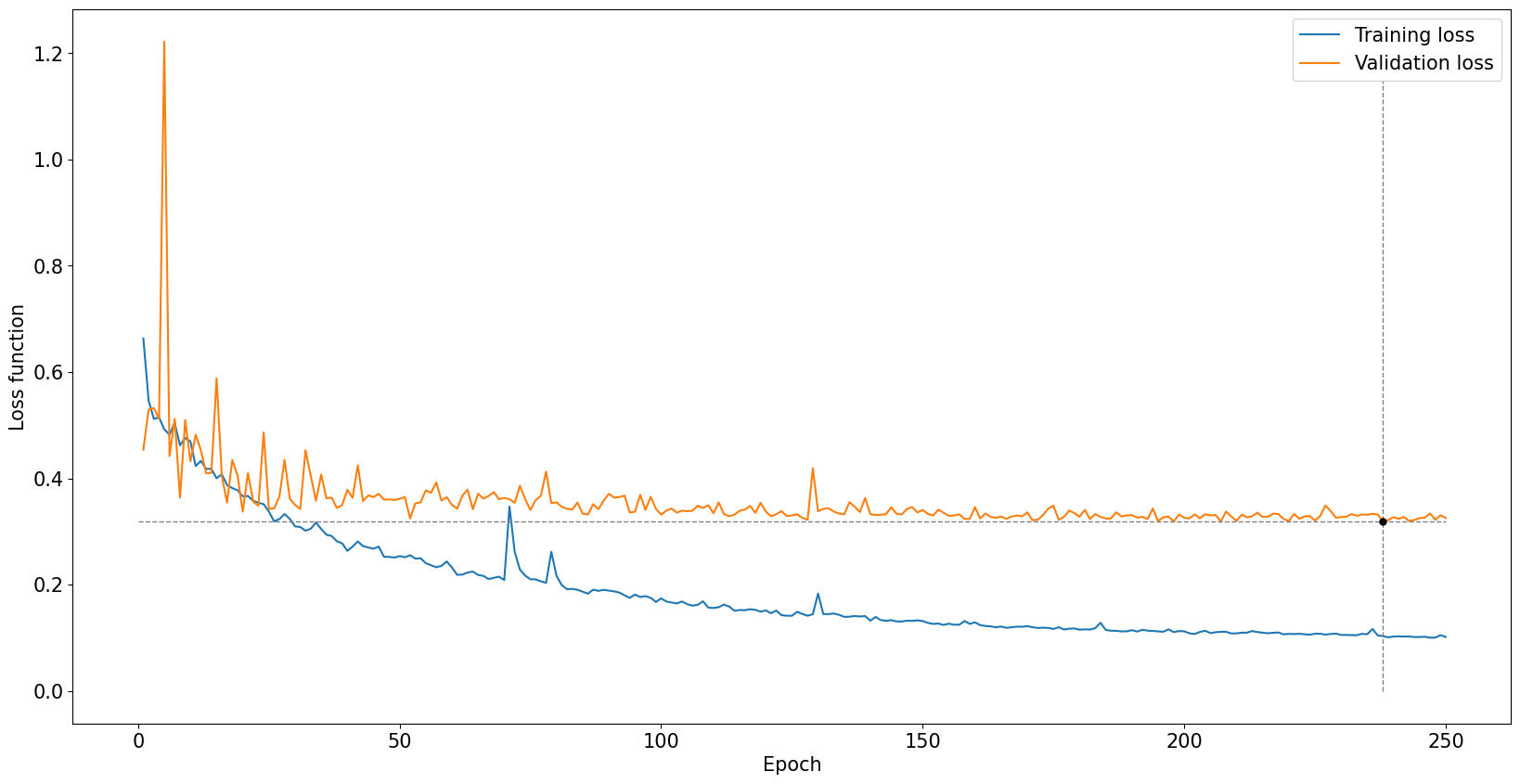}
\caption{Training and validation progress of the CNN model. The grey horizontal and vertical dotted lines indicate the lowest value of the loss function in the validation dataset throughout the epochs and its epoch, respectively. The intersection of the dotted lines indicates the loss function of the validation dataset at epoch 238.}
\label{figure_result_training}
\end{figure*}

\begin{figure*}[!t]
\centering
\includegraphics[width=0.8\textwidth]{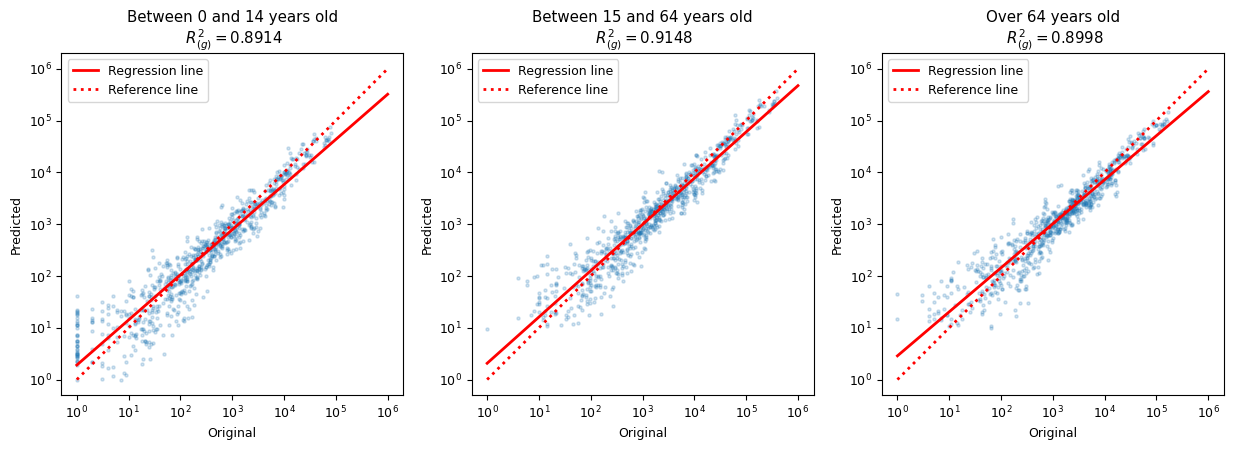}
\caption{Prediction performance of the CNN model in the testing dataset by demographic composition groups. The real lines were drawn from the linear regression of the predicted values on the original ones. The dotted lines were drawn as references which indicate the symmetric normal lines which could have been realised if all the samples would have been predicted precisely.}
\label{figure_result_testing}
\end{figure*}

As a result of the training and validation process, Figure~\ref{figure_result_training} illustrates the training and validation loss function over 250 epochs. In the early epochs, the validation loss function had a volatility of up to 1.2220. While the training loss steadily decreased throughout the epochs, validation loss reached its lower bound of 0.35 at around epoch 50 and little improvement was observed in the following epochs. The training loss recorded 0.3186 as the minima at epoch 238, and the model at the epoch was employed as the one with the most generalisability for the testing and dataset generation.

\begin{table}[h]
\centering
\renewcommand{\arraystretch}{1.3}
\caption{Loss function and $R^2_{(g)}$ at epoch 238 by datasets.}
\label{table_r2_score}

\begin{tabular}{cc|ccc}
\hline
\multicolumn{2}{c|}{Dataset}                                                        & Training & Validation & Testing \\ \hline
\multicolumn{2}{c|}{Loss function}                                                  & 0.1032   & 0.3186     & 0.3215  \\ \cline{1-2}
\multicolumn{1}{c|}{\multirow{3}{*}{$R^2_{(g)}$}} & $g=\text{Btw. 0-14 years old}$  & 0.9963   & 0.8965     & 0.8914  \\ \cline{2-2}
\multicolumn{1}{c|}{}                             & $g=\text{Btw. 15-64 years old}$ & 0.9963   & 0.9173     & 0.9148  \\ \cline{2-2}
\multicolumn{1}{c|}{}                             & $g=\text{Over 64 years old}$        & 0.9959   & 0.8967     & 0.8998  \\ \hline
\end{tabular}

\end{table}

\begin{figure*}[t]
\centering
\includegraphics[width=\textwidth]{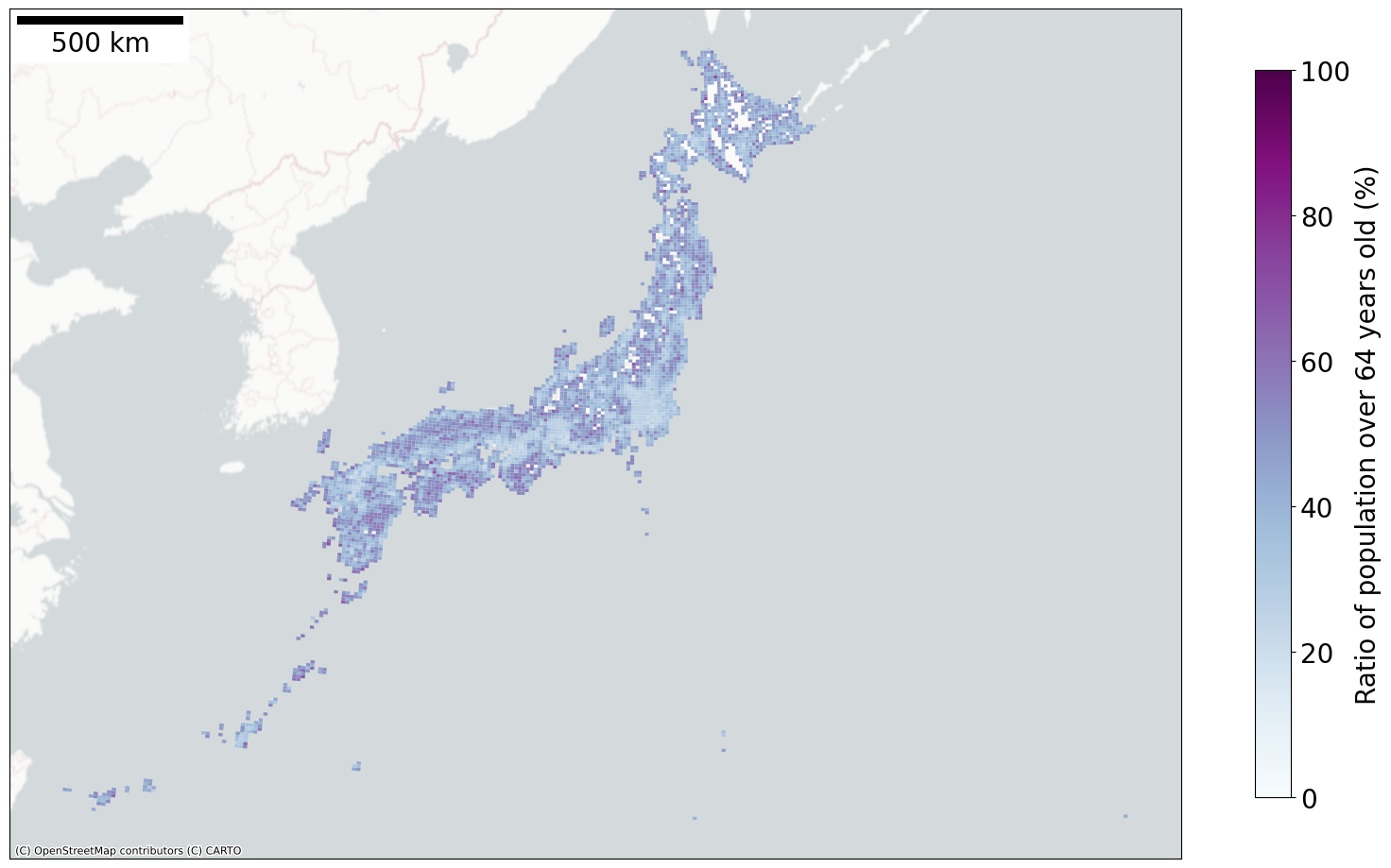}
\caption{Estimated rate of aged population in Japan as of 2022. JGD2011 was applied for the coordinate reference system. The map is based on OpenStreetMap.}
\label{figure_2022}
\end{figure*}

With the trained model chosen above, the testing dataset was utilised for calculating the testing score in $R^2_{(g)}$, as well as the loss function. To do so, all the images in the testing dataset were passed through the CNN model as forward propagation, and the outputs were used for calculating the loss function as a test error. As seen in Table~\ref{table_summary}, the loss function in the testing dataset was 0.3215. This value was close to 0.3186 in the validation dataset, while 0.1032 in the training dataset was relatively smaller than the other two. The outputs obtained were matched with the labels in the testing dataset and $R^2_{(g)}$ was calculated for all the demographic composition groups $g$. As Table~\ref{table_r2_score} and Figure~\ref{figure_result_testing} illustrate, the $R^2_{(g)}$ reached at least 0.8914 for all the demographic composition groups. Although mesh cells with a high population had slight overestimation, the regression lines were aligned along the diagonal of references. Based on the trained CNN model, Figure~\ref{figure_2022} shows the estimated ratio of the population over 64 years old in 2022, which is not a census year. The mesh cells were employed from those where the population was counted either in 2015 or 2020.

\section{Discussion}
As population aging is in progress among an increasing number of countries and regions, prompt understanding of demographic composition is of urgent importance. This paper demonstrated the mesh-wise estimation of demographic composition in Japan as one of the countries with the most aged society in the world, from the dataset of population census in 2015 and 2020 as labels and satellite images of Landsat-8/OLI and Suomi NPP/VIIRS-DNS. The chosen ResNet50-based CNN model after the training and validation process was tested by the testing dataset with metrics of $R^2_{(g)}$ between original and estimated values for each of the demographic composition groups. In the end, the trained model was used for estimating the demographic composition of Japan in 2022 as a non-census year, and the aged population rate was visualised. The result showed a remarkable test score of $R^2_{(g)}$ with at least 0.8914 for all the demographic composition groups, indicating three contributions of following; Firstly, the result displays the effective usage of pre-trained ResNet50 \cite{He_Zhang_Ren_Sun_2016} from the ImageNet dataset consisting of more than 1.2 million images as a 1,000-objects classification task for the demographic composition estimation task with only 6,724 training samples. This result corresponds to previous studies (i.e. \cite{Huang_Zhu_Zhang_Liu_Li_Zou_2021}) in the sense that the remote sensing tasks share a similar domain with the ImageNet object classification task as image recognition. This performance was seen even with the introduction of multi-head inputs in the first convolution layer, where multiple RGB-colour images were accommodated. Secondly, this paper proposes the feasibility of estimating demographic composition rather than total population from satellite images, so that the progress of population aging could be observed without conducting formal surveys such as census. As the values of the loss function both for the validation and testing dataset were not deviated from around 0.32, it is reasonable to confirm that the model held generalisability enough to estimate other geographical locations which were not included in this dataset. Lastly, the proposed CNN model only employed publicly available satellite images as inputs for enduring generalisability in any other geographic location. Since Landsat-8/OLI and Suomi NPP/VIIRS-DNS cover the surface of the earth holistically and periodically, the trained CNN model could be further utilised in other countries in any year. Considering the performance and data availability, the proposed model could help fulfilling the gap between census years not only in Japan but also the other countries and regions for the purposes of urban planning policymaking, especially where rigorous population surveys are limited due to a lack of financial and human resources.

However, there are certain limitations in the proposed CNN model when it comes to enduring generalisability for further usage. First and foremost, biases in the dataset could result in overfitting in other datasets out of this paper. In terms of the spatial biases stemming from the dataset, the proposed CNN model was trained only by the satellite images in the realm of the population census held by the Japanese government. Since this paper did not conduct testing by a dataset from other countries and regions, the generalisability of the trained CNN model is still uncertain if the condition of morphological landscape in satellite images is different from the one in Japan. To consider the temporal biases in the proposed CNN model, the dataset was obtained from the years 2015 and 2020, where each sample was independently treated and did not explicitly consider the temporal relationship in the same geographical location. Also, the satellite images were pre-processed by annually averaging pixel values from those without clouds, which lacked temporal transition within a sample year. From the model architecture point of view, one way to enhance the proposed CNN model for overcoming these problems is to implement spatio-temporal CNN with the introduction of Long Short-Term Memory (LSTM) blocks \cite{Yu_Wang_Liang_Sun_2022} or Attention mechanism \cite{Yang_Yang_Lu_Yang_Liu_Zhou_Fan_2022}. Moreover, rather simply and naturally, data argumentation by expanding the geographical locations and sample years could improve the proposed CNN model. Another limitation is the precision of the proposed CNN model. Although the test score reached 0.8914 for the $R^2_{(g)}$ of all the demographic composition groups $g$ in the testing dataset, Figure~\ref{figure_result_testing} indicates “heteroscedasty” in the correlation between the original and estimated values. In particular, samples with small population tended to be overestimated with poor precision and higher variance than the others. This could be due to the spatial resolution of the satellite images in the input, where 30m for Landasat-8/OLI and 750m for Suomi NPP/VIIRS-DNS were not granular enough to detect the shape of individual buildings in the extent. Not surprisingly, replacement of Landasat-8/OLI and Suomi NPP/VIIRS-DNS with finer satellite images and stacking of other data sources like spatially interpolated POI data would solve this problem. However, this approach would limit the data availability for geographical locations where the data source cannot be obtained. One potential method without employing new data sources is to synthesise satellite images with finer spatial resolution using Generative Adversarial Network (GAN) for super-resolution \cite{Kong_Ryu_Jeong_Zhong_Choi_Kim_Lee_Lim_Jang_Chun_et_al_2023}. In the future study, the proposed CNN model could be applied to other countries and regions with super-resolution to enhance the generalisability and precision for further trustworthy estimation.

\section{Conclusion}
This paper proposed a multi-head CNN model based on pre-trained ResNet50 to estimate demographic composition in Japan as one of the most aged societies from the dataset of satellite images Landsat-8/OLI as daytime and Suomi NPP/VIIRS-DNS as night-time in the year 2015 and 2020. The trained model showed at least 0.8914 in $R^2_{(g)}$ for all the demographic composition groups $g$ consisting of people below 14 years old, between 15 and 64 years old, and over 64 years old in the testing dataset. The trained model estimated the demographic composition in 2022 as a fulfillment of a dataset in one of the non-census years of Japan. Since the proposed CNN model requires inputs only from publicly available satellite images at any location on the earth, it holds generalisability to other countries and regions in terms of data availability. Although further studies are expected for evaluating the performance in different geographical locations and enhancing the coarse spatial resolution in the inputs, this paper shall contribute to policymaking for solving problems amongst population aging without conducting 
a rigorous population survey such as a census.

\section*{Acknowledgment}
The author would like to thank Dr. Elisabetta Pietrostefani as his supervisor in his Master's program at the London School of Economics and Political Science for her advice provision to this paper as the author's dissertation, including the data collection and visualisation methods.


\ifCLASSOPTIONcaptionsoff
  \newpage
\fi


\bibliographystyle{IEEEtran}
\bibliography{IEEEabrv,bibtex/bib/ref}

\begin{thebibliography}{10}
\providecommand{\url}[1]{#1}
\csname url@samestyle\endcsname
\providecommand{\newblock}{\relax}
\providecommand{\bibinfo}[2]{#2}
\providecommand{\BIBentrySTDinterwordspacing}{\spaceskip=0pt\relax}
\providecommand{\BIBentryALTinterwordstretchfactor}{4}
\providecommand{\BIBentryALTinterwordspacing}{\spaceskip=\fontdimen2\font plus
\BIBentryALTinterwordstretchfactor\fontdimen3\font minus
  \fontdimen4\font\relax}
\providecommand{\BIBforeignlanguage}[2]{{%
\expandafter\ifx\csname l@#1\endcsname\relax
\typeout{** WARNING: IEEEtran.bst: No hyphenation pattern has been}%
\typeout{** loaded for the language `#1'. Using the pattern for}%
\typeout{** the default language instead.}%
\else
\language=\csname l@#1\endcsname
\fi
#2}}
\providecommand{\BIBdecl}{\relax}
\BIBdecl

\bibitem{WHO_2022}
\BIBentryALTinterwordspacing
{WHO}, ``Ageing and health,'' 2022, accessed: Aug., 2023. [Online]. Available:
  \url{https://www.who.int/news-room/fact-sheets/detail/ageing-and-health}
\BIBentrySTDinterwordspacing

\bibitem{Bongaarts_2004}
J.~Bongaarts, ``Population {A}ging and the {R}ising {C}ost of {P}ublic
  {P}ensions,'' \emph{Popul. Dev. Rev.}, vol.~30, no.~1, p. 1–23, 2004.

\bibitem{Alley_Liebig_Pynoos_Banerjee_Choi_2007}
D.~Alley, P.~Liebig, J.~Pynoos, T.~Banerjee, and I.~H. Choi, ``Creating
  elder-friendly communities,'' \emph{J. Gerontol. Soc. Work.}, vol.~49, no.
  1–2, p. 1–18, 2007.

\bibitem{Kertzer_White_Bernardi_Gabrielli_2009}
D.~I. Kertzer, M.~J. White, L.~Bernardi, and G.~Gabrielli, ``Italy’s {P}ath
  to {V}ery {L}ow fertility: {T}he {A}dequacy of {E}conomic and {S}econd
  {D}emographic {T}ransition {T}heories,'' \emph{Eur. J. Popul.}, vol.~25,
  no.~1, p. 89–115, Feb 2009.

\bibitem{Census_Gov_2022}
\BIBentryALTinterwordspacing
{US Census Bureau}, ``Decennial census by decade,'' 2022, accessed: Aug., 2023.
  [Online]. Available:
  \url{https://www.census.gov/programs-surveys/decennial-census/decade.html}
\BIBentrySTDinterwordspacing

\bibitem{Office_for_National_Statistics_2023}
\BIBentryALTinterwordspacing
{Office for National Statistics}, ``Census,'' 2023, accessed: Aug., 2023.
  [Online]. Available: \url{https://www.ons.gov.uk/census}
\BIBentrySTDinterwordspacing

\bibitem{World_Bank_2023}
\BIBentryALTinterwordspacing
{World Bank}, ``Population ages 65 and above (
  accessed: Aug., 2023. [Online]. Available:
  \url{https://data.worldbank.org/indicator/SP.POP.65UP.TO.ZS}
\BIBentrySTDinterwordspacing

\bibitem{MLITT_2012}
\BIBentryALTinterwordspacing
{Ministry of Land, Infrastructure, Transport and Tourism}, ``{WHITE PAPER ON
  LAND, INFRASTRUCTURE, TRANSPORT AND TOURISM IN JAPAN, 2012},'' 2012,
  accessed: Aug., 2023. [Online]. Available:
  \url{https://www.mlit.go.jp/english/white-paper/2012.pdf}
\BIBentrySTDinterwordspacing

\bibitem{Feldhoff_2012}
T.~Feldhoff, ``Shrinking communities in {J}apan: Community ownership of assets
  as a development potential for rural japan?'' \emph{URBAN Des. Int.},
  vol.~18, no.~1, p. 99–109, 2012.

\bibitem{Nakatani_2019}
H.~Nakatani, ``Population aging in {J}apan: policy transformation, sustainable
  development goals, universal health coverage, and social determinates of
  health,'' \emph{Glob. Health Med.}, vol.~1, no.~1, p. 3–10, 2019.

\bibitem{Statistics_Bureau_of_Japan_2023b}
\BIBentryALTinterwordspacing
{Statistics Bureau of Japan}, ``population census,'' 2023, accessed: Aug.,
  2023. [Online]. Available:
  \url{https://www.stat.go.jp/english/data/kokusei/index.html}
\BIBentrySTDinterwordspacing

\bibitem{Singleton_Arribas‐Bel_2019}
A.~Singleton and D.~Arribas‐Bel, ``Geographic {D}ata {S}cience,'' \emph{\em
  Geogr. Anal.}, vol.~53, no.~1, p. 61–75, 2019.

\bibitem{Campos-Taberner_García-Haro_Martínez_Izquierdo-Verdiguier_Atzberger_Camps-Valls_Gilabert_2020}
M.~Campos-Taberner, F.~J. García-Haro, B.~Martínez, E.~Izquierdo-Verdiguier,
  C.~Atzberger, G.~Camps-Valls, and M.~A. Gilabert, ``Understanding deep
  learning in land use classification based on {S}entinel-2 time series,''
  \emph{Sci. Rep.}, vol.~10, no.~1, 2020.

\bibitem{USGS_2023a}
\BIBentryALTinterwordspacing
{USGS}, ``{USGS} {M}issions,'' 2023, accessed: Aug., 2023. [Online]. Available:
  \url{https://www.usgs.gov/landsat-missions}
\BIBentrySTDinterwordspacing

\bibitem{ESA_2023}
\BIBentryALTinterwordspacing
{ESA}, ``{SENTINEL-2 MISSION GUIDE},'' 2023, accessed: Aug., 2023. [Online].
  Available: \url{https://sentinel.esa.int/web/sentinel/missions/sentinel-2}
\BIBentrySTDinterwordspacing

\bibitem{NOAA_2023}
\BIBentryALTinterwordspacing
{NOAA}, ``{Suomi NPP Visible Infrared Imaging Radiometer Suite (VIIRS)},''
  2023, accessed: Aug., 2023. [Online]. Available:
  \url{https://ncc.nesdis.noaa.gov/VIIRS}
\BIBentrySTDinterwordspacing

\bibitem{Hu_Patel_Robert_Novosad_Asher_Tang_Burke_Lobell_Ermon_2019}
W.~Hu, J.~H. Patel, Z.-A. Robert, P.~Novosad, S.~Asher, Z.~Tang, M.~Burke,
  D.~Lobell, and S.~Ermon, ``{Mapping Missing Population in Rural India: A Deep
  Learning Approach with Satellite Imagery},'' in \emph{Proceedings of the 2019
  AAAI/ACM Conference on AI, Ethics, and Society}, Honolulu, {HI}, {USA}, Jan.
  2019.

\bibitem{Cheng_Wang_Feng_Yan_2021}
L.~Cheng, L.~Wang, R.~Feng, and J.~Yan, ``{Remote Sensing and Social Sensing
  Data Fusion for Fine-Resolution Population Mapping With a Multimodel Neural
  Network},'' \emph{IEEE J. Sel. Top. Appl. Earth Obs. Remote Sens.}, vol.~14,
  p. 5973–5987, 2021.

\bibitem{Huang_Zhu_Zhang_Liu_Li_Zou_2021}
X.~Huang, D.~Zhu, F.~Zhang, T.~Liu, X.~Li, and L.~Zou, ``{Sensing Population
  Distribution from Satellite Imagery Via Deep Learning:Model Selection,
  Neighboring Effects, and Systematic Biases},'' \emph{IEEE J. Sel. Top. Appl.
  Earth Obs. Remote Sens.}, vol.~14, p. 5137–5151, 2021.

\bibitem{Jenice_Raimond_2016}
A.~R. Jenice and K.~Raimond, ``{An Overview of Technological Revolution in
  Satellite Image Analysis},'' \emph{J. Eng. Sci. Technol. Rev.}, vol.~9,
  no.~4, p. 1–5, 2016.

\bibitem{Phiri_Simwanda_Salekin_Nyirenda_Murayama_Ranagalage_2020}
D.~Phiri, M.~Simwanda, S.~Salekin, V.~Nyirenda, Y.~Murayama, and M.~Ranagalage,
  ``{Sentinel-2 Data for Land Cover/Use Mapping: A Review},'' \emph{Remote
  Sens.}, vol.~12, no.~14, p. 2291, 2020.

\bibitem{Wu_Murray_2007}
C.~Wu and A.~T. Murray, ``{Population Estimation Using Landsat Enhanced
  Thematic Mapper Imagery},'' \emph{Geogr. Anal.}, vol.~39, no.~1, p. 26–43,
  2007.

\bibitem{Barentine_Walczak_Gyuk_Tarr_Longcore_2021}
J.~C. Barentine, K.~Walczak, G.~Gyuk, C.~Tarr, and T.~Longcore, ``{A Case for a
  New Satellite Mission for Remote Sensing of Night Lights},'' \emph{Remote
  Sens.}, vol.~13, no.~12, p. 2294, 2021.

\bibitem{Henderson_Yeh_Gong_Elvidge_Baugh_2003}
M.~Henderson, E.~T. Yeh, P.~Gong, C.~Elvidge, and K.~Baugh, ``Validation of
  urban boundaries derived from global night-time satellite imagery,''
  \emph{Int. J. Remote Sens.}, vol.~24, no.~3, p. 595–609, 2003.

\bibitem{Henderson_Storeygard_Weil_2012}
J.~V. Henderson, A.~Storeygard, and D.~N. Weil, ``{Measuring Economic Growth
  from Outer Space},'' \emph{Am. Econ. Rev.}, vol. 102, no.~2, p. 994–1028,
  2012.

\bibitem{Mellander_Lobo_Stolarick_Matheson_2015}
C.~Mellander, J.~Lobo, K.~Stolarick, and Z.~Matheson, ``{Night-Time Light Data:
  A Good Proxy Measure for Economic Activity?}'' \emph{PLOS ONE}, vol.~10,
  no.~10, 2015.

\bibitem{Bagan_Yamagata_2015}
H.~Bagan and Y.~Yamagata, ``Analysis of urban growth and estimating population
  density using satellite images of nighttime lights and land-use and
  population data,'' \emph{GIsci. Remote Sens.}, vol.~52, no.~6, p. 765–780,
  2015.

\bibitem{Wang_Huang_Zhao_Hou_Zhang_Gu_2020}
Y.~Wang, C.~Huang, M.~Zhao, J.~Hou, Y.~Zhang, and J.~Gu, ``{Mapping the
  Population Density in Mainland China Using NPP/VIIRS and Points-Of-Interest
  Data Based on a Random Forests Model},'' \emph{Remote Sens.}, vol.~12,
  no.~21, p. 3645, 2020.

\bibitem{Rumelhart_Hinton_Williams_1986}
D.~E. Rumelhart, G.~E. Hinton, and R.~J. Williams, ``Learning representations
  by back-propagating errors,'' \emph{Nature}, vol. 323, no. 6088, p.
  533–536, 1986.

\bibitem{Deng_2014}
L.~Deng, ``A tutorial survey of architectures, algorithms, and applications for
  deep learning,'' \emph{APSIPA Trans. Signal Inf. Process.}, vol.~3, no.~1,
  2014.

\bibitem{Krizhevsky_Sutskever_Hinton_2017}
A.~Krizhevsky, I.~Sutskever, and G.~E. Hinton, ``{ImageNet Classification with
  Deep Convolutional Neural Networks},'' \emph{Commun. ACM}, vol.~60, no.~6, p.
  84–90, 2017.

\bibitem{Szegedy_Wei_Liu_Yangqing_Jia_Sermanet_Reed_Anguelov_Erhan_Vanhoucke_Rabinovich_2015}
C.~Szegedy, W.~Liu, Y.~Jia, P.~Sermanet, S.~Reed, D.~Anguelov, D.~Erhan,
  V.~Vanhoucke, and A.~Rabinovich, ``{Going Deeper with Convolutions},'' in
  \emph{2015 IEEE Conference on Computer Vision and Pattern Recognition
  (CVPR)}, Boston, {MA}, {USA}, Jun. 2015, pp. 1--9.

\bibitem{Simonyan_Zisserman_2015}
\BIBentryALTinterwordspacing
K.~Simonyan and A.~Zisserman, ``Very deep convolutional networks for
  large-scale image recognition,'' 2015. [Online]. Available:
  \url{https://arxiv.org/abs/1409.1556}
\BIBentrySTDinterwordspacing

\bibitem{He_Zhang_Ren_Sun_2016}
C.~Szegedy, W.~Liu, Y.~Jia, P.~Sermanet, S.~Reed, D.~Anguelov, D.~Erhan,
  V.~Vanhoucke, and A.~Rabinovich, ``{Deep Residual Learning for Image
  Recognition},'' in \emph{2016 IEEE Conference on Computer Vision and Pattern
  Recognition (CVPR)}, Las {V}egas, {NV}, {USA}, Jun. 2016.

\bibitem{Russakovsky_Deng_Su_Krause_Satheesh_Ma_Huang_Karpathy_Khosla_Bernstein_et_al_2015}
O.~Russakovsky, J.~Deng, H.~Su, J.~Krause, S.~Satheesh, S.~Ma, Z.~Huang,
  A.~Karpathy, A.~Khosla, M.~Bernstein, and et~al., ``Imagenet large scale
  visual recognition challenge,'' \emph{Int. J. Comput. Vis.}, vol. 115, no.~3,
  p. 211–252, 2015.

\bibitem{Zhuang_Qi_Duan_Xi_Zhu_Zhu_Xiong_He_2021}
F.~Zhuang, Z.~Qi, K.~Duan, D.~Xi, Y.~Zhu, H.~Zhu, H.~Xiong, and Q.~He, ``A
  comprehensive survey on transfer learning,'' \emph{Proc. IEEE}, vol. 109,
  no.~1, p. 43–76, 2021.

\bibitem{Shin_Roth_Gao_Lu_Xu_Nogues_Yao_Mollura_Summers_2016}
H.-C. Shin, H.~R. Roth, M.~Gao, L.~Lu, Z.~Xu, I.~Nogues, J.~Yao, D.~Mollura,
  and R.~M. Summers, ``{Deep Convolutional Neural Networks for Computer-Aided
  Detection: CNN Architectures, Dataset Characteristics and Transfer
  Learning},'' \emph{IEEE Trans. Med. Imaging}, vol.~35, no.~5, p. 1285–1298,
  2016.

\bibitem{Yang_Guo_Yue_Liu_Hu_Li_2019}
J.~Yang, J.~Guo, H.~Yue, Z.~Liu, H.~Hu, and K.~Li, ``{CDnet: CNN-Based Cloud
  Detection for Remote Sensing Imagery},'' \emph{IEEE Trans. Geosci. Remote
  Sens.}, vol.~57, no.~8, p. 6195–6211, 2019.

\bibitem{de_Bem_de_Carvalho_Junior_Fontes_Guimarães_Trancoso_Gomes_2020}
P.~de~Bem, O.~de~Carvalho~Junior, R.~Fontes~Guimarães, and R.~Trancoso~Gomes,
  ``{Change detection of deforestation in the Brazilian amazon using landsat
  data and Convolutional Neural Networks},'' \emph{Remote Sens.}, vol.~12,
  no.~6, p. 901, 2020.

\bibitem{Ulmas_Liiv_2020}
\BIBentryALTinterwordspacing
P.~Ulmas and I.~Liiv, ``{Segmentation of Satellite Imagery using U-Net Models
  for Land Cover Classification},'' 2020. [Online]. Available:
  \url{https://arxiv.org/abs/2003.02899}
\BIBentrySTDinterwordspacing

\bibitem{Solórzano_Mas_Gao_Gallardo-Cruz_2021}
J.~V. Solórzano, J.~F. Mas, Y.~Gao, and J.~A. Gallardo-Cruz, ``Land use land
  cover classification with u-net: Advantages of combining sentinel-1 and
  sentinel-2 imagery,'' \emph{Remote Sens.}, vol.~13, no.~18, p. 3600, 2021.

\bibitem{USGS_2023b}
\BIBentryALTinterwordspacing
{USGS}, ``{Landsat Collection 2 Level-2 Science Products},'' 2023, accessed:
  Aug., 2023. [Online]. Available:
  \url{https://www.usgs.gov/landsat-missions/landsat-collection-2-level-2-science-products}
\BIBentrySTDinterwordspacing

\bibitem{eStat_2023}
\BIBentryALTinterwordspacing
{e-Stat}, ``{Portal Site of Official Statistics of Japan},'' 2023, accessed:
  Aug., 2023. [Online]. Available:
  \url{{https://www.e-stat.go.jp/en/stat-search?page=1}}
\BIBentrySTDinterwordspacing

\bibitem{Statistics_Bureau_of_Japan_2023a}
\BIBentryALTinterwordspacing
{Statistics Bureau of Japan}, ``{Method of Demarcation for Grid Square},''
  2023, accessed: Aug., 2023. [Online]. Available:
  \url{https://www.stat.go.jp/english/data/mesh/05.html}
\BIBentrySTDinterwordspacing

\bibitem{Kaplan_Avdan_Avdan_2018}
G.~Kaplan, U.~Avdan, and Z.~Y. Avdan, ``{Urban Heat Island Analysis Using the
  Landsat 8 Satellite Data: A Case Study in Skopje, Macedonia},'' in \emph{The
  2nd International Electronic Conference on Remote Sensing}, no.~7, Mar. 2018,
  p. 358.

\bibitem{Özelkan_2020}
E.~Özelkan, ``Water body detection analysis using ndwiindices derived from
  landsat-8 oli,'' \emph{Pol. J. Environ.}, vol.~29, no.~2, p. 1759–1769,
  2020.

\bibitem{Google_2023}
\BIBentryALTinterwordspacing
Google, ``{USGS Landsat 8 Level 2, Collection 2, Tier 1},'' 2023. [Online].
  Available:
  \url{{https://developers.google.com/earth-engine/datasets/catalog/LANDSAT\_LC08\_C02\_T1\_L2}}
\BIBentrySTDinterwordspacing

\bibitem{Elvidge_Zhizhin_Hsu_Baugh_2013}
C.~Elvidge, M.~Zhizhin, F.-C. Hsu, and K.~Baugh, ``{VIIRS Nightfire: Satellite
  Pyrometry at Night},'' \emph{Remote Sens.}, vol.~5, no.~9, p. 4423–4449,
  2013.

\bibitem{Elvidge_Zhizhin_Ghosh_Hsu_Taneja_2021}
C.~D. Elvidge, M.~Zhizhin, T.~Ghosh, F.-C. Hsu, and J.~Taneja, ``{Annual time
  series of global VIIRS Nighttime Lights derived from monthly averages: 2012
  to 2019},'' \emph{Remote Sens.}, vol.~13, no.~5, p. 922, 2021.

\bibitem{Earth_Observation_Group_2023}
\BIBentryALTinterwordspacing
{Earth Observation Group}, ``{VIIRS Nighttime Light},'' 2023. [Online].
  Available: \url{https://eogdata.mines.edu/products/vnl}
\BIBentrySTDinterwordspacing

\bibitem{Kaufman_Rosset_Perlich_2011}
S.~Kaufman, S.~Rosset, and C.~Perlich, ``{Leakage in Data Mining},'' \emph{ACM
  Trans. Knowl. Discov. Data}, vol.~6, no.~4, p. 1–21, 2011.

\bibitem{Paszke_Gross_Massa_Lerer_Bradbury_Chanan_Killeen_Lin_Gimelshein_Antiga_et_al_2019}
\BIBentryALTinterwordspacing
A.~Paszke, S.~Gross, F.~Massa, A.~Lerer, J.~Bradbury, G.~Chanan, T.~Killeen,
  Z.~Lin, N.~Gimelshein, L.~Antiga, and et~al., ``{PyTorch: An Imperative
  Style, High-Performance Deep Learning Library},'' 2019. [Online]. Available:
  \url{https://arxiv.org/abs/1912.01703}
\BIBentrySTDinterwordspacing

\bibitem{Kingma_Ba_2017}
\BIBentryALTinterwordspacing
D.~P. Kingma and J.~Ba, ``{Adam: A Method for Stochastic Optimization},'' 2017.
  [Online]. Available: \url{https://arxiv.org/abs/1412.6980}
\BIBentrySTDinterwordspacing

\bibitem{Ioffe_Szegedy_2015}
\BIBentryALTinterwordspacing
S.~Ioffe and C.~Szegedy, ``{Batch Normalization: Accelerating Deep Network
  Training by Reducing Internal Covariate Shift},'' 2015. [Online]. Available:
  \url{https://arxiv.org/abs/1502.03167}
\BIBentrySTDinterwordspacing

\bibitem{Li_Chen_Hu_Yang_2019}
X.~Li, S.~Chen, X.~Hu, and J.~Yang, ``{Understanding the Disharmony between
  Dropout and Batch Normalization by Variance Shift},'' in \emph{2019 IEEE/CVF
  Conference on Computer Vision and Pattern Recognition (CVPR)}, {L}ong
  {B}each, {CA}, {USA}, Mar. 2019.

\bibitem{Amazon_2023}
\BIBentryALTinterwordspacing
{AWS}, ``{Amazon EC2 G4 Instances},'' 2023. [Online]. Available:
  \url{https://aws.amazon.com/ec2/instance-types/g4/}
\BIBentrySTDinterwordspacing

\bibitem{Yu_Wang_Liang_Sun_2022}
D.~Yu, X.~Wang, P.~Liang, and X.~Sun, ``Spatio-temporal convolutional residual
  network for regional commercial vitality prediction,'' \emph{Multimed. Tools
  Appl.}, vol.~81, no.~19, p. 27923–27948, 2022.

\bibitem{Yang_Yang_Lu_Yang_Liu_Zhou_Fan_2022}
G.~Yang, Y.~Yang, Z.~Lu, J.~Yang, D.~Liu, C.~Zhou, and Z.~Fan, ``{STA-TSN:
  Spatial-Temporal Attention Temporal Segment Network for action recognition in
  video},'' \emph{PLOS ONE}, vol.~17, no.~3, 2022.

\bibitem{Kong_Ryu_Jeong_Zhong_Choi_Kim_Lee_Lim_Jang_Chun_et_al_2023}
J.~Kong, Y.~Ryu, S.~Jeong, Z.~Zhong, W.~Choi, J.~Kim, K.~Lee, J.~Lim, K.~Jang,
  J.~Chun, and et~al., ``{Super resolution of historic landsat imagery using a
  dual generative Adversarial Network (GAN) model with CubeSat constellation
  imagery for spatially enhanced long-term vegetation monitoring},''
  \emph{ISPRS J. Photogramm. Remote Sens.}, vol. 200, p. 1–23, 2023.

\end{thebibliography}

\end{document}